\documentclass[11pt,twocolumn]{article}

\usepackage[margin=0.75in]{geometry}
\usepackage{graphicx}
\usepackage{amsmath, amssymb}
\usepackage{booktabs}
\usepackage{microtype}
\usepackage[numbers]{natbib}
\usepackage{xurl}

\usepackage{titlesec}
\usepackage{enumitem}
\usepackage{balance}
\usepackage{flushend}

\titlespacing*{\section}{0pt}{0.9\baselineskip}{0.55\baselineskip}
\titlespacing*{\subsection}{0pt}{0.7\baselineskip}{0.35\baselineskip}

\setlist[itemize]{leftmargin=*, itemsep=2pt, topsep=3pt, parsep=0pt, partopsep=0pt}

\usepackage{hyperref}
\hypersetup{
  colorlinks=true,
  linkcolor=black,
  citecolor=black,
  urlcolor=blue
}

\title{
Natural Language Declarative Prompting (NLD-P):\\
A Modular Governance Method for Prompt Design Under Model Drift
}

\author{
Hyunwoo Kim \hspace{1.2em} Hanau Yi \hspace{1.2em} Jaehee Bae \hspace{1.2em} Yumin Kim \\
ddai Inc. \\
\texttt{\{hw.kim, hnu.yi, jh.bae, ym.kim\}@ai-dda.com}
}

\date{}

\begin{document}

\twocolumn[
\maketitle
\vspace{-0.75em}

\begin{abstract}
\noindent
The rapid evolution of large language models (LLMs) has transformed prompt engineering from a localized craft into a systems-level governance challenge. As models scale and update across generations, prompt behavior becomes sensitive to shifts in instruction-following policies, alignment regimes, and decoding strategies, a phenomenon we characterize as GPT-scale model drift. This paper reconceptualizes Natural Language Declarative Prompting (NLD-P) as a declarative governance method rather than a rigid field template. NLD-P is formalized as a modular control abstraction that separates provenance, constraint logic, task content, and post-generation evaluation, encoded directly in natural language without reliance on orchestration code. Portions of drafting employed a schema-bound LLM assistant configured under NLD-P. All methodological claims were directed and verified under a documented human-in-the-loop protocol. We conclude by outlining implications for declarative control under ongoing model evolution and directions for future validation.
\end{abstract}
\vspace{1.2\baselineskip}
]

\section{Introduction: Prompting as Governance Under Model Drift}

Large language models (LLMs) have evolved from relatively static generative systems into continuously updated architectures shaped by shifting alignment policies, decoding strategies, and instruction-following heuristics. Post-training alignment procedures such as reinforcement learning from human feedback (RLHF) demonstrate that substantial behavioral changes can occur without modifying base model architecture (Ouyang et al., 2022). Foundation model research further situates model behavior within a socio-technical ecosystem in which training data, deployment context, and governance regimes co-produce system outputs (Bommasani et al., 2021). From a sociotechnical perspective, technical artifacts embed forms of governance and constraint within their design (Winner, 1980). Contemporary system documentation confirms that successive model releases incorporate iterative safety mitigations, alignment refinements, and revised refusal policies, reinforcing that instruction-following characteristics are governed across versions rather than fixed at pretraining (Singh et al., 2025).

At the interface level, these governance dynamics are increasingly mediated through natural language. Linguistic analysis indicates that sustained interaction with LLM systems gives rise to a hybridized register shaped by system affordances and alignment constraints (Kim \& Yi, 2025). Termed \emph{machine-facing English}, this register reflects the co-adaptation of human expression to model behavior and model alignment to recurring prompting patterns (Kim \& Yi, 2025). As model generations introduce behavioral adjustments through alignment and deployment updates, prompts once treated as stable artifacts become sensitive to architectural change. Under these conditions, prompt engineering must be reconceptualized not as phrasing craft but as governance enacted within an evolving linguistic interface (Kim \& Yi, 2025).

\subsection{The Escalation of Prompt Control}

Early prompt practices relied on role assignment, iterative rephrasing, or example-based steering. These approaches implicitly assumed behavioral continuity across model updates. Empirical findings demonstrate, however, that scaling and alignment tuning significantly influence reasoning style, verbosity, and adherence to declared constraints (Ouyang et al., 2022; Wei et al., 2022). Even minor variations in prompt framing or example ordering can alter performance characteristics (Min et al., 2022). Frontier model documentation further shows that refusal behavior and constraint handling are adjusted between releases as part of iterative deployment processes (Singh et al., 2025).

Simultaneously, linguistic evidence suggests that users adapt prompting strategies in response to perceived system sensitivities, gradually internalizing alignment-aware phrasing patterns (Kim \& Yi, 2025). This co-adaptive dynamic intensifies the control burden on language itself. As models evolve through post-training and policy updates, identical prompts may vary in constraint enforcement, structural formatting, or interpretive emphasis. Such variability reflects architectural and governance change rather than user inconsistency.

Consequently, prompt design shifts from stylistic optimization to structural control. Effective prompting increasingly requires explicit constraint visibility and modular logic that can be revised without rewriting entire instruction blocks. Without such separation, behavioral drift accumulates and undermines reproducibility across model versions. From a governance perspective, prompt authorship resembles reflective practice under conditions of uncertainty, where design and revision form a continuous feedback loop (Schön, 1983). Prompt control becomes structured oversight within a non-stationary technical system rather than rhetorical refinement alone. Because governance is expressed linguistically, structural clarity must counteract the tendency of machine-facing registers to absorb constraints into diffuse narrative phrasing (Kim \& Yi, 2025).

\subsection{GPT-Scale Model Drift}

We define \emph{GPT-scale model drift} as cross-version variability in instruction-following behavior resulting from architectural updates, alignment modifications, or inference-mode differences. Unlike classical distribution shift, which arises from changes in input data distributions (Quionero-Candela et al., 2009), GPT-scale drift stems from revisions to model parameters or safety constraints between releases. Concept drift literature addresses non-stationary environments in which predictive relationships evolve over time (Gama et al., 2014); GPT-scale drift differs insofar as internal policy updates are intentionally introduced through model revision cycles. Public documentation confirms that alignment refinements and safety policies are revised across releases, introducing version-dependent variation in constraint handling (Singh et al., 2025).

Drift may manifest as inconsistent enforcement of declared constraints, reinterpretation of task boundaries, changes in verbosity, or altered adherence to post-generation conditions. Safety research has emphasized that objective misalignment and distributional fragility can arise even in high-performing systems (Amodei et al., 2016). From a systems perspective, any shift in constraint behavior introduces instability into externally authored control logic.

When governance is encoded in natural language, drift interacts with register-level adaptation. As models shift through updated alignment regimes, users may adjust phrasing to compensate, reinforcing the co-evolution of linguistic form and system behavior (Kim \& Yi, 2025). Drift therefore operates at both architectural and interface levels, where governance and language co-adapt. The central challenge lies in designing control abstractions that remain interpretable and revisable under architectural change rather than assuming behavioral continuity.

\subsection{The Governance Gap in Natural-Language Interfaces}

Developer-centric AI pipelines externalize control logic into code, enabling structured orchestration and validation. Such approaches support reproducibility but require technical infrastructure and programming fluency. In contrast, many LLM users operate exclusively through natural-language interfaces, where governance must be encoded linguistically within the prompt itself.

Research in end-user programming highlights the difficulty non-developers face when abstract control structures are implicit rather than tool-supported (Ko et al., 2011). Human–AI interaction guidelines similarly emphasize transparency and user control as prerequisites for trustworthy AI systems (Amershi et al., 2019). Concurrently, linguistic analysis indicates that as users adapt to evolving model behavior, prompting increasingly reflects system-oriented phrasing patterns that prioritize alignment compatibility over conversational naturalness (Kim \& Yi, 2025).

When governance logic is conflated with task instructions in a single narrative block, users must manage control implicitly within this evolving register. This increases fragility under model change. Conventional prompting often merges task objectives, behavioral constraints, formatting rules, and evaluation expectations into undifferentiated text. Such conflation obscures governance logic and complicates targeted revision when drift occurs. Accountability frameworks in AI governance advocate lifecycle-based oversight rather than post hoc remediation (Raji et al., 2020). In natural-language interfaces, however, such lifecycle governance must itself be expressed in language already shaped by model-mediated conventions (Kim \& Yi, 2025). The burden of structural clarity thus shifts decisively to prompt design.

\subsection{From Format to Method}

Natural Language Declarative Prompting (NLD-P) was initially introduced through structured field-based prompts separating identity, rules, and task content. Over time, durability proved to depend not on specific bracket syntax but on a deeper invariant: explicit separation between provenance, constraint logic, task execution, and post-generation evaluation.

This structural separation functions as a countermeasure to linguistic drift within machine-facing registers. If governance remains embedded in diffuse narrative phrasing, it becomes vulnerable to reinterpretation under evolving alignment heuristics and mitigation regimes. By contrast, modular segmentation preserves governance clarity even when the surrounding register shifts.

This principle aligns with modularity theory in software architecture, where separation of concerns enhances system stability and maintainability (Parnas, 1972; Shaw \& Garlan, 1996). In complex systems theory, hierarchical decomposition allows local modification without systemic collapse (Simon, 1962). By externalizing governance layers from execution logic, modular systems enable targeted revision rather than cascading restructuring.

This paper formalizes NLD-P as a declarative governance method rather than a formatting convention. Its core contribution lies in the externalization of control logic into modular, legible natural-language components that remain interpretable under model drift. The objective is not to eliminate drift but to render it visible and tractable. When governance layers are structurally separated from task content, behavioral deviations can be addressed through localized revision of constraint blocks rather than wholesale prompt reconstruction.

The following sections formalize the invariants of declarative modular prompting, define minimal compliance criteria, and analyze schema receptivity across evolving LLM architectures. NLD-P is positioned not as a performance optimization technique but as a governance abstraction designed for structural stability under GPT-scale model drift and linguistic co-adaptation in human–AI discourse (Kim \& Yi, 2025).

\section{Declarative Governance Without Code}

NLD-P is formalized as a declarative governance method for prompt construction under conditions of model drift. Its central innovation lies in modular separation of instruction layers within natural language itself. Governance logic is made explicit, inspectable, and revision-ready without reliance on external scripting, orchestration frameworks, or weight modification. Whereas developer-centric pipelines externalize control into code, NLD-P embeds governance directly within linguistic structure.

This formulation draws conceptually from separation-of-concerns principles in software architecture, where isolating functional components enhances system stability and maintainability (Parnas, 1972; Shaw \& Garlan, 1996). It also parallels the distinction between declarative and procedural paradigms in programming theory, in which constraints are articulated as explicit conditions rather than embedded within execution flow (Dijkstra, 1976). In NLD-P, modularity is achieved not computationally but linguistically: structure is enacted through segmentation of control and task layers within natural-language prompts.

\subsection{Core Invariants of NLD-P}

NLD-P rests on four structural invariants. A prompt aligns with the method when these elements are explicitly separated and independently interpretable:

\begin{itemize}
\item \textbf{Provenance}: Operational context, role assumptions, or scope conditions relevant to execution.
\item \textbf{Constraint Logic}: Behavioral rules, formatting requirements, or boundary conditions declared independently of the task objective.
\item \textbf{Task Content}: The primary objective isolated from governance directives.
\item \textbf{Post-Generation Evaluation}: Conditions for validation, revision, or compliance articulated prior to execution.
\end{itemize}

These invariants externalize governance. Rather than embedding expectations implicitly within a single narrative instruction, NLD-P distinguishes between what the model should do and how its behavior should be constrained or evaluated. This structural separation mirrors modular design theory, where isolating components reduces cascading instability when one component changes (Parnas, 1972). By disentangling execution logic from governance logic, prompts become structurally inspectable, diagnosable, and locally revisable under changing model conditions.

\subsection{Schema Versus Surface Syntax}

Early implementations of NLD-P employed visible field markers to reinforce structural separation. However, the durability of the method does not depend on specific labels, brackets, or markup conventions. The essential requirement is structural modularity rather than syntactic uniformity.

A prompt conforms to NLD-P when constraint logic remains distinct from task content, governance instructions are independently revisable, and evaluation conditions are declared rather than implied. Surface syntax may vary across platforms, interfaces, or stylistic conventions; structural separation, by contrast, preserves governance clarity under behavioral change. As architectural theory emphasizes, robustness derives from conceptual layering rather than syntactic representation (Shaw \& Garlan, 1996). NLD-P is therefore defined at the level of structural invariants rather than template rigidity.

\subsection{Minimal Compliance Criteria}

For analytical clarity, a prompt is considered NLD-P compliant if it:

\begin{itemize}
\item Explicitly separates control logic from task instructions.
\item Encodes behavioral constraints in natural language prior to execution.
\item Articulates post-generation validation or revision conditions.
\item Supports targeted revision without collapsing governance and task layers.
\end{itemize}

These criteria are descriptive rather than prescriptive. They define alignment with the method without requiring adherence to a single canonical schema. Compliance is architectural rather than syntactic: what matters is the persistence of modular boundaries rather than their formatting.

\subsection{Reference Schema Illustration}

To clarify how these invariants may be instantiated, Table~\ref{tab:nldp_layers} presents the abstract structural layers of a canonical NLD-P configuration. The representation is illustrative rather than normative. Block names function as authoring conventions and are not machine-parsed metadata. Compliance depends on preservation of structural separation rather than specific surface labels.

\begin{table*}[t]
\centering
\begin{tabular}{lll}
\toprule
\textbf{Block} & \textbf{Functional Role} & \textbf{Example Elements} \\
\midrule
IDENT & Provenance and execution context & schema, assistant, mode, type, version \\
RULE & Constraint logic layer & tone, clarity flags, formatting constraints \\
CONTENT & Task execution layer & Primary objective or instruction \\
INPUT & Optional contextual grounding & Reference text or external material \\
POSTCHECK & Post-generation evaluation & Validation rules, revision triggers \\
\bottomrule
\end{tabular}
\caption{Canonical NLD-P Structural Layers}
\label{tab:nldp_layers}
\end{table*}

A compact schematic instantiation is shown in Figure~\ref{fig:nldp_example}. This illustration demonstrates structural separation rather than prescriptive syntax.

\begin{figure*}[t]
\centering
\begin{minipage}{0.9\textwidth}
\begin{verbatim}
[IDENT]
schema = NLD-P
assistant = Evalyn
mode = essay_proofreader
type = declarative_guidance
version = 2.0

[RULE:OUTPUT_STYLE]
tone = formal
style = academic
enforce_clarity = true

[CONTENT]
Proofread the paragraph and identify ambiguity.

[POSTCHECK]
If vague modal verbs are detected, flag for revision.
If citation format deviates from APA, suggest correction.
\end{verbatim}
\end{minipage}
\caption{Illustrative Canonical NLD-P Instantiation}
\label{fig:nldp_example}
\end{figure*}

Layered separation encodes governance logic directly within natural language. The model does not interpret these labels as executable metadata; their function is authorial transparency and lifecycle traceability. The abstraction resides in the persistence of modular boundaries rather than in specific markers.

\subsection{Declarative Modularity as Control Abstraction}

Traditional developer-centric pipelines externalize governance through code, validation scripts, or orchestration frameworks. NLD-P instead achieves modularity within the prompt itself. While external tooling may enhance scalability, the core governance abstraction remains embedded in natural language.

Declarative modularity therefore functions as a control abstraction. By isolating governance components, prompts become structurally transparent and revision-ready. When behavior shifts, authors can adjust constraint blocks without rewriting task logic. This localized adjustability reflects classical modular system design, in which abstraction enables local modification without systemic collapse (Parnas, 1972; Shaw \& Garlan, 1996).

The significance of NLD-P lies not in template rigidity but in abstraction of governance into legible components. Under GPT-scale model drift, explicit structural separation provides a stable conceptual scaffold even as model behavior evolves.

\section{Prompt Lifecycle as Structured Governance}

Under model drift, prompt design must be treated as an iterative governance cycle rather than a one-time instruction. NLD-P structures this cycle into four stages: authoring, execution, evaluation, and revision. Each stage remains modular, enabling targeted adjustment without collapsing the overall architecture. This lifecycle orientation aligns with models of reflective practice in which design, action, and reassessment form a continuous loop under conditions of uncertainty (Schön, 1983). It also parallels double-loop learning frameworks, where underlying governing variables—not merely surface behaviors—are examined and revised (Argyris \& Schön, 1978). Rather than viewing prompting as a static artifact, NLD-P conceptualizes it as managed governance embedded within evolving system behavior.

\subsection{Authoring with Constraint Visibility}

Authoring begins with explicit declaration of control logic prior to execution. Behavioral boundaries, formatting requirements, and evaluation conditions are articulated independently of the task objective. This separation clarifies intent before interpretation and preserves editability: constraints may be strengthened, relaxed, or reformulated without modifying core task logic. Constraint visibility therefore functions as pre-execution governance.

From a systems perspective, making control conditions explicit increases interpretability and reduces ambiguity in downstream evaluation. Human–AI interaction research identifies transparency and user control as prerequisites for reliable oversight in AI-mediated systems (Amershi et al., 2019). By isolating constraints in advance, authoring becomes structured boundary-setting rather than post hoc correction. The prompt shifts from a monolithic instruction to a layered governance specification.

\subsection{Execution Under Drift Conditions}

Execution unfolds within model environments that vary across versions, deployment contexts, or inference modes. Even when instructions remain unchanged, interpretation of constraints may shift due to alignment updates or scaling effects (Ouyang et al., 2022; Wei et al., 2022). In monolithic prompts, such shifts are difficult to diagnose because task instructions and governance logic are intertwined.

Modular separation enables behavioral variation to be traced to specific control layers. If adherence weakens, revision can target the constraint boundary rather than the task narrative. Execution is thus reframed not as an opaque generation event but as a structural stress test of governance clarity. This framing aligns with lifecycle accountability approaches in AI governance, where system behavior is evaluated relative to explicitly declared policies rather than implicit expectations (Raji et al., 2020). Drift becomes analyzable rather than anecdotal.

\subsection{Declarative Evaluation}

Evaluation is defined prior to execution. The prompt may declare explicit conditions governing acceptable outputs, including formatting requirements, prohibited categories, structural constraints, or verification rules. By articulating evaluation criteria ex ante, NLD-P embeds oversight within the design stage itself.

When violations occur, corrective logic can be revised directly within the governance layer. Evaluation becomes intrinsic to the lifecycle rather than appended externally. This approach parallels documentation-based accountability frameworks emphasizing advance specification of evaluation standards and transparent reporting practices (Mitchell et al., 2019). Declarative evaluation transforms prompt refinement from reactive adjustment into structured compliance review. Rather than revising outputs ad hoc, authors revise governance specifications.

\subsection{Revision and Stability}

Revision targets discrete governance components. Because provenance, constraints, task content, and evaluation logic are structurally separated, iteration does not require full prompt reconstruction. Adjustments can be localized to constraint blocks, evaluation criteria, or contextual assumptions without destabilizing the task layer.

This targeted revision supports stability under drift. The method does not eliminate variability but reduces cascading instability produced by conflated instruction structures. Stability emerges from preservation of governance boundaries rather than assumption of invariant model behavior. Lifecycle governance therefore transforms prompting from reactive trial-and-error into structured iteration. Continuity resides in architectural separation, even as model behavior evolves.

\section{Schema Receptivity and Model Dependency}

Declarative modular prompting operates within model architectures that differ in how they interpret instruction layers. Even when governance logic is made explicit, its effectiveness depends on how a model parses constraint boundaries relative to task content. Research on large language model scaling and behavioral predictability shows that responses shift systematically with training scale and alignment configuration (Ganguli et al., 2022). Studies of alignment regimes further demonstrate that constitutional and reinforcement-based approaches yield materially different patterns of constraint adherence and refusal behavior (Anthropic, 2023). Prompt sensitivity research indicates that structural scaffolding and reasoning decomposition can substantially alter outputs even when task content remains constant (Yao et al., 2022). System-level documentation across frontier deployments further illustrates that mitigation layering and refusal policies evolve over time, introducing version-dependent variation in constraint enforcement.

This interaction between structural prompt design and model-specific interpretation introduces what we term \emph{schema receptivity}: the degree to which a given model recognizes, respects, and preserves declared governance separations. Schema receptivity is not binary but gradient, shaped by training distributions, alignment objectives, safety filters, moderation pipelines, and decoding heuristics. Governance logic expressed in natural language is therefore mediated through evolving alignment regimes rather than fixed syntactic parsing.

\subsection{Model-Dependent Constraint Adherence}

Different LLM generations exhibit varying sensitivities to declared constraints. Some models adhere closely to explicit formatting or behavioral boundaries, whereas others reinterpret them through broader alignment heuristics or safety policies. Empirical red-teaming research demonstrates that modest variations in prompt framing can expose substantial differences in safety enforcement and boundary interpretation (Perez et al., 2022). Experimental work on in-context learning further shows that demonstration structure and example ordering significantly influence adherence patterns (Min et al., 2022).

Alignment methodology itself affects constraint behavior. Constitutional alignment approaches modify how models internalize normative guidelines relative to alternative post-training methods (Anthropic, 2023). Such differences indicate that constraint adherence is mediated not only by base architecture but also by post-training governance layers.

This variability does not invalidate declarative governance. Rather, it underscores that instruction-following capacity is neither static nor homogeneous across model iterations. Schema receptivity acknowledges that governance abstractions operate within evolving policy regimes. Declarative modularity does not assume invariant compliance; instead, it provides structural visibility when compliance fluctuates. When adherence weakens or shifts, modular separation allows authors to reinforce, clarify, or recalibrate governance blocks without altering task content.

\subsection{Stabilization Through Explicit Control Surfaces}

Declarative constraint blocks function as identifiable control surfaces. They provide discrete loci for adjustment when behavior shifts across model versions. Instead of embedding expectations diffusely within narrative phrasing, governance directives remain isolated, inspectable, and revisable. This reflects modular system design principles in which isolating components reduces propagation of unintended side effects during modification (Parnas, 1972).

In contemporary deployment practice, alignment refinements and mitigation updates are layered at both model and system levels. Explicit governance blocks within prompts operate analogously at the interface layer, enabling targeted revision when alignment regimes shift without requiring wholesale reconstruction of task instructions. Empirical analyses of scaling and predictability suggest that even when aggregate capabilities stabilize, constraint interpretation may continue to vary at the interaction layer (Ganguli et al., 2022).

Stabilization in this context does not imply invariant outputs across model generations. Rather, it reflects increased responsiveness to structured revision. When constraints are explicitly articulated, deviations can be localized and addressed systematically. Drift becomes diagnosable within prompt architecture rather than obscured within entangled instructions. Schema receptivity therefore determines not whether governance universally succeeds, but whether structural separation remains interpretable and actionable under changing model conditions.

\subsection{Portability and Deployment Context}

NLD-P is defined at the level of architectural abstraction rather than platform-specific tooling. Its principles apply across hosted interfaces, API environments, and research workflows. Governance logic remains embedded in natural language rather than external orchestration frameworks. Interface-level governance is increasingly recognized as a primary site of control in foundation model ecosystems (Bommasani et al., 2021).

Portability is conceptual rather than syntactic. Structural separation preserves interpretability even as deployment contexts, inference modes, or model versions change. Constraint adherence may vary according to schema receptivity, yet the governance abstraction itself remains intact. By maintaining explicit boundaries between task execution and behavioral control, NLD-P supports cross-context interpretability under architectural evolution and model dependency.

\section{Reference Implementation and Transparency}

The formalization of NLD-P emerged through sustained application in drafting, analytical refinement, and governance-oriented prompt design workflows. Because large language models can influence structure, phrasing, and argumentative flow, explicit delineation of system assistance and authorship boundaries is required. Transparency in documentation and attribution has become central in responsible AI research, particularly in relation to model reporting, dataset documentation, and disclosure of system capabilities (Mitchell et al., 2019; Gebru et al., 2021). Broader critiques of large language models further emphasize epistemic responsibility and the risks of obscuring system mediation in knowledge production (Bender et al., 2021). Emerging scholarly discourse on generative AI and authorship similarly stresses that accountability for intellectual claims must remain with human contributors.

This section clarifies the operational role of the LLM assistant and the limits of its contribution within the present manuscript. Declarative modular prompting functions not only as a design method but also as a transparency instrument, rendering the boundary between governance logic and execution visible within authorship practice.

\subsection{Evalyn as a Schema-Bound Drafting Agent}

An LLM assistant configured under the NLD-P framework was used as a structured drafting instrument. It operated under explicitly declared governance constraints, including role definitions, citation discipline, structural limits, and stylistic parameters embedded within the prompt schema. These constraints were articulated prior to generation and functioned as boundary conditions governing output scope.

The assistant did not introduce independent research claims nor autonomously define conceptual frameworks or theoretical positions. All substantive claims were determined by the human author. Generated drafts were treated as provisional artifacts subject to review, revision, or rejection. The system functioned as a reference implementation of declarative modular prompting rather than as an experimental subject or intellectual collaborator. Its role was to execute schema-bound instructions within a human-directed governance structure.

\subsection{Human-in-the-Loop Finalization}

All conceptual framing, definitional boundaries, and scope limitations were determined by the human author. Draft outputs were reviewed iteratively, edited for conceptual precision, and revised to align with the intended theoretical structure. No section was included without explicit approval and refinement. Oversight was continuous rather than post hoc, reflecting lifecycle governance principles consistent with accountability frameworks in AI system development (Raji et al., 2020).

Constraint blocks were themselves revised across drafting cycles to maintain clarity and alignment with architectural claims. Human supervision governed both content selection and structural modification. The LLM functioned as a drafting instrument within a controlled declarative environment rather than as a source of autonomous intellectual contribution.

\subsection{Authorship and Responsibility}

The intellectual content of this manuscript remains the responsibility of the human author. LLM-assisted drafting does not alter authorship attribution or epistemic accountability. Transparent disclosure of system involvement aligns with documentation practices such as model cards and dataset reporting standards emphasizing clarity of responsibility and system capabilities (Mitchell et al., 2019; Gebru et al., 2021).

By explicitly defining the assistant’s operational scope and maintaining continuous human oversight, the implementation preserves methodological integrity while avoiding ambiguity regarding intellectual ownership. Declarative governance thus functions not only as a prompt design abstraction but also as a transparency mechanism, ensuring that authorship, accountability, and system assistance remain structurally differentiated.

\section{Limitations and Scope}

This paper formalizes NLD-P as a declarative governance method under GPT-scale model drift. It does not claim empirical superiority, performance gains, or universal stability across model architectures. The contribution is architectural rather than benchmark-driven, situating NLD-P within broader discussions of alignment-sensitive interface design and foundation-model governance (Bommasani et al., 2021). Contemporary research demonstrates that alignment configurations and post-training regimes materially influence model behavior (Ganguli et al., 2022; Anthropic, 2023). NLD-P addresses this dynamism at the level of interface abstraction rather than model internals.

No quantitative benchmarking is provided. The method is articulated at the level of structural abstraction rather than experimental comparison. Systematic measurement of constraint adherence across model generations remains future work. Rigorous validation would require controlled cross-version testing, clearly defined compliance metrics, and replication across deployment environments sensitive to evolving alignment regimes and inference configurations.

Declarative modularity does not eliminate variability. Model-dependent schema receptivity implies that constraint adherence may differ across contexts, inference modes, safety layers, or alignment configurations. NLD-P increases visibility and revisability of governance logic but does not guarantee invariant execution. Stability derives from architectural separation rather than behavioral uniformity.

The reference implementation reflects contemporary deployment environments and current-generation instruction-following models. Operational nuances may vary across future architectures, particularly as alignment methods, tool-augmented reasoning, retrieval integration, or multi-agent orchestration frameworks become more prevalent. As foundation models evolve and governance layers grow more complex (Bommasani et al., 2021), interface-level abstractions may require adaptation to remain interpretable and effective.

Minimal compliance criteria are descriptive rather than exhaustive. They define alignment with the method at the level of structural separation without prescribing canonical syntax. Alternative frameworks may satisfy similar invariants under different terminology or representational conventions. NLD-P should therefore be understood as one articulation of declarative modular governance rather than an exclusive formalism.

Future work should include empirical evaluation across model generations, measurement of schema receptivity under controlled perturbations, and comparative analysis with programmatic orchestration approaches, including tool-based pipelines and hybrid declarative–procedural systems. Longitudinal studies examining how linguistic adaptation interacts with evolving alignment regimes would further clarify the relationship between architectural drift and interface-level governance (Kim \& Yi, 2025). The present contribution addresses the architectural question of how governance can be externalized within natural language under model drift, leaving empirical validation to subsequent research.

\section{Conclusion}

As large language models continue to evolve, prompt engineering must be reframed as a governance problem embedded within an adaptive technical and linguistic system. Architectural updates, alignment refinements, and deployment shifts alter how constraints are interpreted over time. When governance is enacted through natural language, these shifts interact with the evolving register of human–AI discourse (Kim \& Yi, 2025). Surface formatting conventions and monolithic instruction blocks are therefore insufficient to preserve interpretability under conditions of model drift.

This paper formalizes Natural Language Declarative Prompting (NLD-P) as a declarative modular governance method. Its defining property is explicit structural separation between provenance, constraint logic, task content, and post-generation evaluation. By externalizing control layers into independently revisable components, NLD-P renders drift visible rather than opaque. Governance becomes inspectable and locally adjustable at the interface level, even as alignment regimes and model architectures evolve.

NLD-P does not guarantee invariant behavior across systems. Its contribution is architectural: it articulates a control abstraction whose coherence does not depend on behavioral uniformity. Structural separation enables targeted revision without collapsing task logic, allowing governance to adapt alongside evolving models while maintaining conceptual continuity.

As users adapt language to model sensitivities, machine-facing registers may absorb governance into diffuse narrative phrasing (Kim \& Yi, 2025). Declarative modularity counteracts this tendency by preserving the distinction between execution and control. It sustains structural clarity at the intersection of linguistic adaptation and architectural change.

The central claim of this work is therefore architectural rather than empirical: when governance is expressed through natural language, modular separation becomes a design requirement. Under GPT-scale model drift and linguistic co-adaptation, explicit governance segmentation is not stylistic preference but structural necessity for preserving interpretability and revisability over time.

...
\end{document}